\documentclass[11pt,a4paper]{article}
\usepackage[hyperref]{acl2021}
\usepackage{times}
\usepackage{amsmath}
\usepackage{multirow}
\usepackage{url}
\usepackage{float}
\usepackage{pgfplots}
\usepackage{footnote}
\usepackage[T5]{fontenc}
\usepackage[utf8]{inputenc}

\usepackage{wasysym}
\usepackage{amssymb}

\usepackage{makecell}
\usepackage{rotating}

\usepackage{multicol}
\usepackage{array}
\usepackage{flushend}

\usepackage{makecell}

\usepackage{flafter}

\usepackage{graphicx}
\graphicspath{ {./images/} }

\aclfinalcopy



\title{SMTCE: A Social Media Text Classification Evaluation Benchmark\\and BERTology Models for Vietnamese}

\author{
    Luan Thanh Nguyen$^{1, 2}$, 
    Kiet Van Nguyen$^{1, 2}$,
    \bf Ngan Luu-Thuy Nguyen$^{1, 2}$ \\
    $^{1}$Faculty of Information Science and Engineering, University of Information Technology, \\Ho Chi Minh City, Vietnam \\
    $^{2}$Vietnam National University, Ho Chi Minh City, Vietnam \\
    \texttt{\{luannt, kietnv, ngannlt\}@uit.edu.vn}
    }

\date{}
  
\date{}
\begin{document}
\maketitle

\begin{abstract} 

Text classification is a typical natural language processing or computational linguistics task with various interesting applications. As the number of users on social media platforms increases, data acceleration promotes emerging studies on \textbf{S}ocial \textbf{M}edia \textbf{T}ext \textbf{C}lassification (\textbf{SMTC}) or social media text mining on these valuable resources. In contrast to English, Vietnamese, one of the low-resource languages, is still not concentrated on and exploited thoroughly. Inspired by the success of the GLUE, we introduce the \textbf{S}ocial \textbf{M}edia \textbf{T}ext \textbf{C}lassification \textbf{E}valuation (\textbf{SMTCE}) benchmark, as a collection of datasets and models across a diverse set of SMTC tasks. With the proposed benchmark, we implement and analyze the effectiveness of a variety of multilingual BERT-based models (mBERT, XLM-R, and DistilmBERT) and monolingual BERT-based models (PhoBERT, viBERT, vELECTRA, and viBERT4news) for tasks in the SMTCE benchmark. Monolingual models outperform multilingual models and achieve state-of-the-art results on all text classification tasks. It provides an objective assessment of multilingual and monolingual BERT-based models on the benchmarks, which will benefit future studies about BERTology in the Vietnamese language. 

\end{abstract}

\section{Introduction}

With the rise of social media, researching user comments may allow us to understand their behavior and enhance the quality of cyberspace. Social text classification is one of the current popular NLP tasks that aim to solve that problem with an extensive study on comments users left on social media platforms. There are several different current tasks in social media text classification (SMTC) tasks in several domains, and however, it is designed to be independent of other inconsistent data domains.

Motivated by the popularity of the GLUE \cite{wang-etal-2018-glue}, we present a novel Social Media Text Classification Evaluation (SMTCE) benchmark with four social media text classification tasks in Vietnamese, including constructive speech detection \cite{nguyen2021constructive}, complaint comment dectection \cite{nguyen2021vietnamese}, emotion recognition \cite{uit-vsmec}, and hate speech detection \cite{luu2021largescale} tasks.

Natural Language Processing (NLP), a subset of artificial intelligence techniques, is rapidly evolving, with numerous notable accomplishments. The primary goal of assisting computers in understanding human language is the bridge for reducing the distance between humans and computers. The current trend in NLP is releasing language models that have been pre-trained on a considerable amount of data to perform various tasks. Hence, we do not need to code from scratch for every task and annotate millions of texts each time. This is the premise for transfer learning and applying the architectures transformers to various downstream tasks in NLP.

Bidirectional Encoder Representations from Transformers, known as BERT, have appeared and played an essential role in the current outstanding development of computational linguistics. BERT has become a typical baseline in NLP experiments. Several numerous kinds of research have been conducted to analyze the performance of BERTology models \cite{rogers-etal-2020-primer}. Following its release, multilingual BERT-based models emerged, achieving breakthrough performances in serving several languages. Moreover, monolingual language models are also concentrated, primarily focused on a single language, particularly a low-resource language like Vietnamese. In this study, we conduct experiments with multilingual and monolingual BERT-based language models on the SMTCE benchmark to see how they perform.

Our main contributions to this research are:
\begin{itemize}

    \item We propose \textbf{SMTCE}, the {\textbf S}ocial \textbf{M}edia {\textbf T}ext {\textbf C}lassification {\textbf E}valuation benchmark for evaluating social media text classification or social media mining in Vietnamese.
    
    \item We implement various BERT-based models on the SMTCE benchmark. We then compare and contrast the characteristics and strengths of monolingual and multilingual BERT-based language models on computing Vietnamese text classification tasks.
    
    \item After achieving results of monolingual language models, which outperform multilingual models, we start to discuss the remaining problems that researchers on social media text mining have to face.
    
\end{itemize}

The remainder of this work is structured as follows: Section 2 includes related works to which we refer. Section 3 describes the tasks in the SMTCE benchmark. Section 4 overviews BERT-based language models available for Vietnamese NLP tasks. Section 5 shows the processes for implementing models and their results for each task in the SMTCE benchmark. Section 6 is the discussion of the remaining problems in this study. Section 7, the last section, is our conclusion for this research and future work.

\section{Related Work}

The introduction of BERT, \citet{devlin-etal-2019-bert} led to the explosion of the transformer language models. BERT has obtained state-of-the-art performances on a variety of NLP tasks upon launch, including nine GLUE tasks, SQuAD v1.0 and 2.0, and SWAG. 


Shortly after releasing BERT, \citet{devlin-etal-2019-bert} then published multilingual BERT, which was capable of over 100 languages. Then, a slew of BERTology models started to emerge. \citet{NEURIPS2019_c04c19c2} released XLM, a cross-lingual language model achieving promising results on various NLP tasks. Following the introduction RoBERTa, they introduced a new pre-trained model, XLM-R \cite{conneau-etal-2020-unsupervised}, which reached breakthrough results.

Aside from multilingual versions or BERT-based models, developing NLP in countries with different languages promotes researchers to build and improve monolingual models based on available BERT architectures for their languages. We have CamemBERT \cite{martin-etal-2020-camembert} for France, Chinese-BERT \cite{cui-etal-2020-revisiting} for Chinese, or BERT-Japanese \cite{bertjapanese} for Japanese. One of the low-resource languages, Vietnamese, has monolingual BERT-based models that have been pre-trained on Vietnamese datasets, such as: PhoBERT \cite{nguyen-tuan-nguyen-2020-phobert}, viBERT \cite{bui-etal-2020-improving}, vELECTRA \cite{bui-etal-2020-improving}, or viBERT4news\footnote{https://github.com/bino282/bert4news}. 

In Vietnamese, \citet{to2021monolingual} did research about investigating monolingual and multilingual BERT-based models for the Vietnamese summarization task. It is the first attempt to use datasets from other languages based on multilingual models to execute various existing pre-trained language models on the summarization task.


In this paper, we implement several monolingual and multilingual BERT-based pre-trained language models on the proposed SMTCE benchmark tasks. Ensuingly, we conduct an overview of these two types of models in Vietnamese SMTC tasks.

\begin{table*}
\centering
\caption{Statistics and descriptions of tasks in the SMTCE benchmark. All datasets used the macro-average F1-score to measure the performance of machine learning models.}
\label{tab:sta_des_STCE}
\resizebox{\linewidth}{!}{%
\begin{tabular}{lrrrlrrl} 
\hline
\multicolumn{1}{c}{\textbf{Dataset}} & \multicolumn{1}{c}{\textbf{Train}} & \multicolumn{1}{c}{\textbf{Dev}} & \multicolumn{1}{c}{\textbf{Test}} & \multicolumn{1}{c}{\textbf{Task}} & \multicolumn{1}{c}{\textbf{IAA}} & \multicolumn{1}{c}{\begin{tabular}[c]{@{}c@{}}\textbf{Baseline Result}\\\textbf{ (F1-macro \%)}\end{tabular}} & \multicolumn{1}{c}{\textbf{Data Source}} \\ 
\hline
\multicolumn{8}{c}{\textit{Binary text classification}} \\ 
\hline
ViCTSD & 7,000 & 2,000 & 1,000 & Constructive speech detection & 0.59 & 78.59 & News comments \\
ViOCD & 4,387 & 548 & 549 & Complaint comment detection & 0.87 & 92.16 & E-commerce feedback \\ 
\hline
\multicolumn{8}{c}{\textit{Multi-class text classification}} \\ 
\hline
VSMEC & 5,548 & 686 & 693 & Emotion recognition & 0.80 & 59.74 & Social network comments \\
ViHSD & 24,048 & 2,672 & 6,680 & Hate speech detection & 0.52 & 62.69 & Social network comments \\
\hline
\end{tabular}
}
\end{table*}
\section{Social Media Text Classification Tasks}
Technology is continuously changing, and social networks allow our users to interact and exchange information more easily. Because of this ease, many harmful and malicious comments of anonymous users aimed to attack individuals psychologically. Studies in this field are gaining attraction to help automatically classify comments as helpful, constructive, or harmful to block and hide them promptly. By giving suitable solutions depending on various situations, we hope to create positive and friendly cyberspace. 

In this study, we propose SMTCE, a new benchmark concentrating on four social media text classification tasks, which covers various domains, data sizes, and challenges in social media tasks. Overview of the tasks in the SMTCE is shown in Table \ref{tab:sta_des_STCE} with the statistics and descriptions of tasks in the SMTCE benchmark, including dataset name, the number of texts of each set, task target, inter-annotator agreement (IAA), baseline result, and data source.

\subsection{Emotion Recognition Task}
\begin{figure}[H]
 \centering
 \includegraphics[width=1.\linewidth]{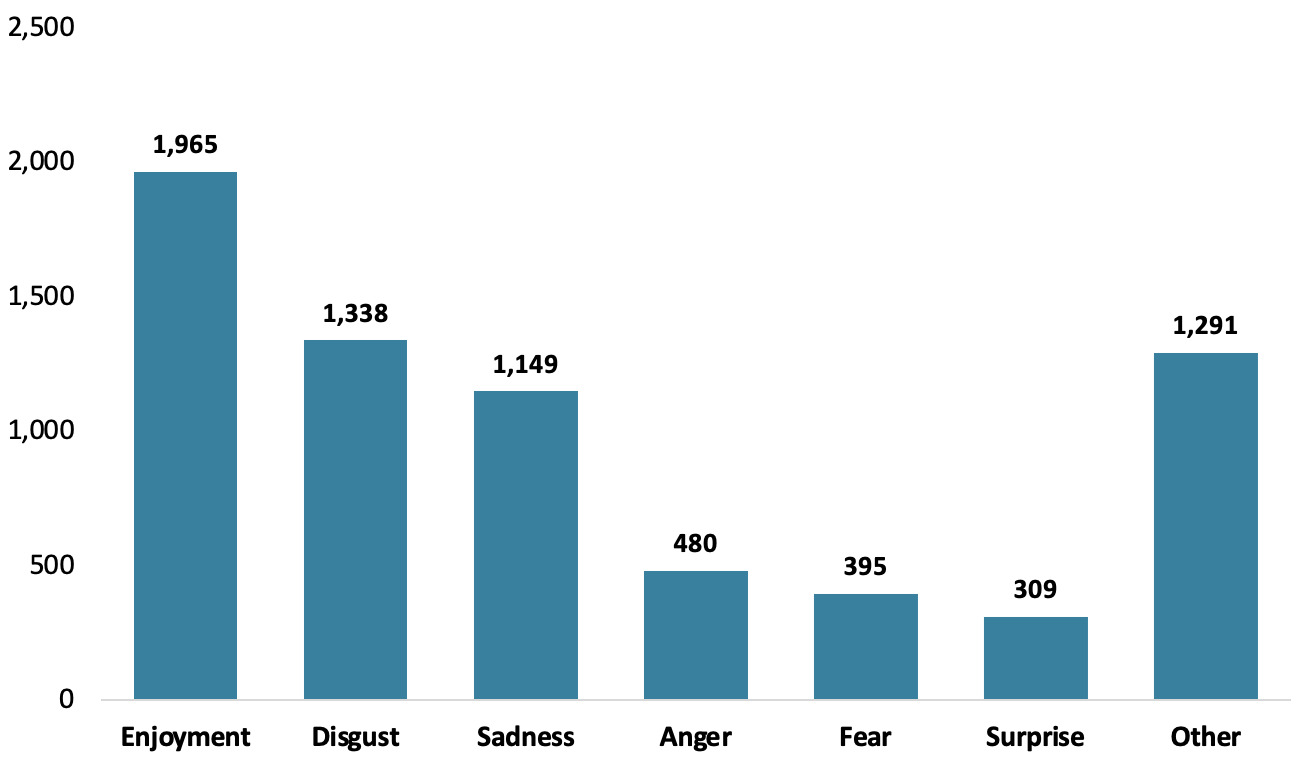}
 \caption{The number of sentences of each label in the VSMEC dataset.}
 \label{fig:VSMEC_Analysis}
\end{figure}

\citet{uit-vsmec} released a standard Vietnamese Social Media Emotion Corpus known as UIT-VSMEC (VSMEC) for solving the task of recognizing the emotion of Vietnamese comments on social media. It is made up of 6,927 sentences that had been manually annotated with seven emotion labels, including anger, disgust, enjoyment, fear, sadness, surprise, and other. Figure \ref{fig:VSMEC_Analysis} illustrates the number of sentences of each label in the dataset. As we can see, the highest number of sentences belongs to the label enjoyment with 1,965 sentences, and the lowest is the surprise label with 309 sentences.

\subsection{Constructive Speech Detection Task}
\begin{figure}[H]
 \centering
 \includegraphics[width=1.\linewidth]{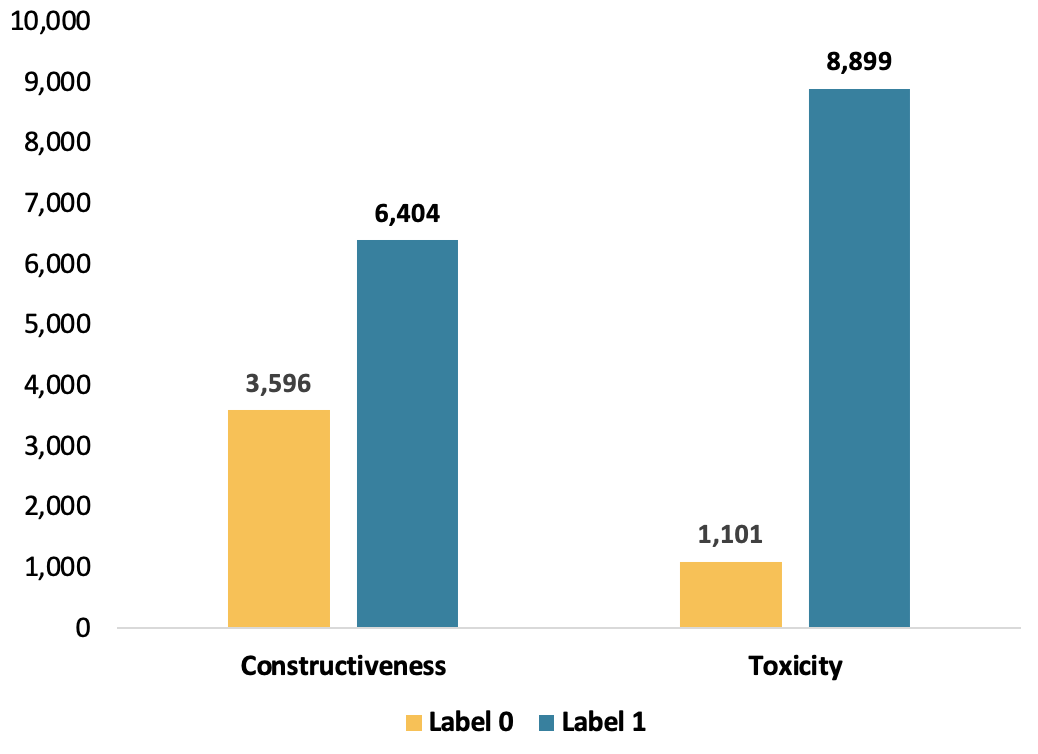}
 \caption{The statistic of the number of sentences of each label in the ViCTSD dataset.}
 \label{fig:ViCTSD_Analysis}
\end{figure}

The dataset UIT-ViCTSD (ViCTSD: Vietnamese Constructive and Toxic Speech Detection) \citet{nguyen2021constructive} was built for dealing with the task of detecting constructive and toxic speech. This task aims to solve two issues: detecting constructive and toxic speech in Vietnamese social media comments. Each comment is labeled into two different tasks: identifying constructive and toxic comments. To define constructiveness, there are two labels: constructive and non-constructive. Furthermore, there are two labels for classifying toxic comments: toxic and non-toxic. Figure \ref{fig:ViCTSD_Analysis} below is the statistic of the number of each label in the dataset.

The dataset contains 10,000 comments by ten domains of users crawled from the online discussion as VnExpress.net. The dataset serves to identify the constructiveness and toxicity of Vietnamese social media comments. The authors also evaluated the first version of this dataset with a proposed system with 78.59\% and 59.40\% F1-score for constructiveness and toxicity detection.

As depicted in Figure \ref{fig:ViCTSD_Analysis}, we see that the dataset is very imbalanced in toxic speech detection tasks, so we only focus on the task of constructiveness detection in this study.

\subsection{Hate Speech Detection Task}

\begin{figure}[H]
 \centering
 \includegraphics[width=1.\linewidth]{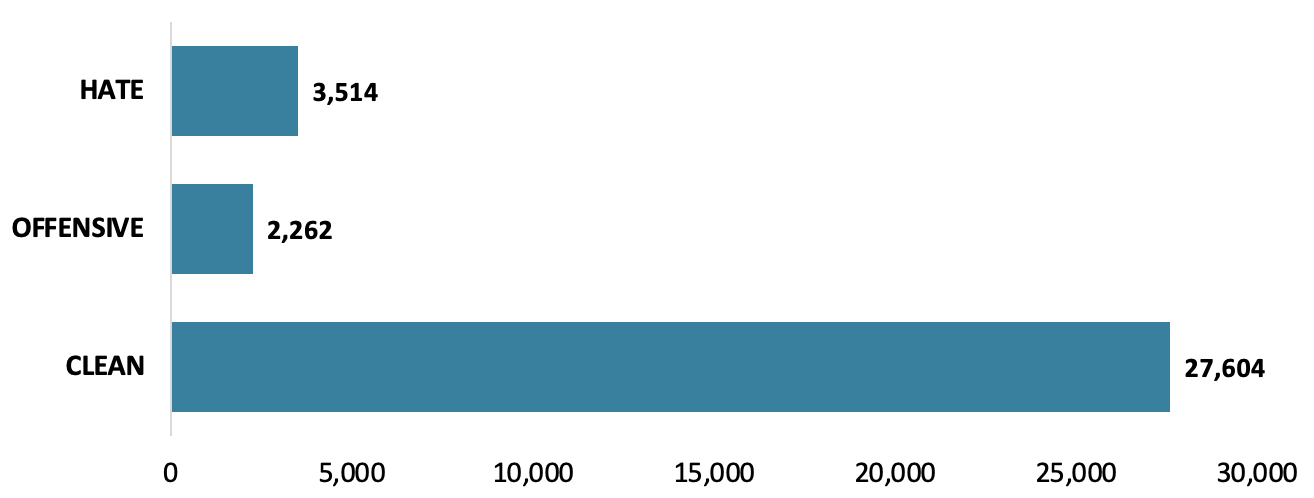}
 \caption{The statistic of the number of sentences of each label in the ViHSD dataset.}
 \label{fig:ViHSD_Analysis}
\end{figure}

\citet{luu2021largescale} provided a dataset for the task of hate speech detection on Vietnamese social media comments named UIT-ViHSD (ViHSD). The dataset includes 30,000 comments labeled by annotators with three labels CLEAN, OFFENSIVE, and HATE. We describe the proportion of each label in the dataset in Figure \ref{fig:ViHSD_Analysis}. As shown in this illustration, this dataset is severely unbalanced (82.71\% of CLEAN comments) with a low inter-annotator agreement of only 0.52, and it required several techniques in pre-processing data to deal with this imbalance. 

\begin{table*}
\centering
\caption{An overview of available BERTology languages model in Vietnamese.}
\label{tab:BERTology}
\begin{tabular}{clllll} 
\hline
\multicolumn{1}{l}{} &  & \multicolumn{1}{c}{\textbf{Data size}} & \multicolumn{1}{c}{\textbf{Vocab. size}} & \multicolumn{1}{c}{\textbf{Tokenization}} & \multicolumn{1}{c}{\textbf{Domain}} \\ 
\hline
\multirow{3}{*}{\begin{tabular}[c]{@{}c@{}}\textbf{Multilingual}\end{tabular}} & mBERT (case  uncased) & 16GB & 3.3B & Subword & Book+Wiki \\
 & XLM-R (base) & 2.5TB & 250K & Subword & Common Crawl \\
 & DistilmBERT & 16GB & 31K & Subword & Book+Wiki \\ 
\hline
\multirow{4}{*}{\begin{tabular}[c]{@{}c@{}}\textbf{Monolingual}\end{tabular}} & PhoBERT & 20GB & 64K & Subword & News+Wiki \\
 & viBERT & 10GB & 32K & Subword & News \\
 & vELECTRA & 60GB & 32K & Subword & News \\
 & viBERT4news & 20GB & 62K & Syllable & News \\
\hline
\end{tabular}
\end{table*}

Hate speech has been a source of concern for social media users. The dataset aims to create a tool for identifying it in online communication interactions, censoring it to protect users from offensive content, and improving the environment of online forums.

\subsection{Complaint Comment Detection Task}
\citet{nguyen2021vietnamese} researched customer complaints on e-commerce sites and released a novel open-domain dataset named UIT-ViOCD (ViOCD), a collection of 5,485 human-annotated comments on four domains. It was then evaluated using multiple approaches and achieving the best performance with an F1-score of 92.16\% by the fine-tuned PhoBERT model. We can utilize this information to classify complaint comments from users on open-domain social media automatically. Table \ref{tab:ViOCD_Analysis} below depicts the distribution of each label in sets. 

\begin{table}[H]
\centering
\caption{The distribution of each label in the train, valid, and test sets of the ViOCD dataset.}
\label{tab:ViOCD_Analysis}
\begin{tabular}{lrr}
\hline
 & \multicolumn{1}{c}{\textbf{Complaint}} & \multicolumn{1}{c}{\textbf{Non-complaint}} \\ \hline
\textbf{Train set} & 2,292          & 2,095          \\
\textbf{Dev set} & 283           & 265           \\
\textbf{Test set}  & 279           & 270           \\ \hline
\textbf{Total}     & \textbf{2,854} & \textbf{2,630} \\ \hline
\end{tabular}
\end{table}

Even though this dataset contains a small number of data points, the ratio between labels 0 and 1 is quite balanced. Hence, in this task, we do not need to use significant techniques to deal with data imbalance before starting the training session.
\section{BERTology Language Models}
Transformer-based models are currently the best-performance models in natural language processing or computational linguistics. The architecture of the transformer model consists of two parts: encoder and decoder. It differs from previous deep learning architectures that it pushes in all of the input data at once rather than progressively, thanks to the self-attention \cite{NIPS2017_3f5ee243} mechanism. Following the premiere of \citet{devlin-etal-2019-bert}, models based on BERT architectures are becoming increasingly popular. Challenging NLP tasks have been effectively solved using these models.

BERT (Bidirectional Encoder Representations from Transformers), the same architecture as the bi-directional recurrent neural network, uses bi-directional encoder representations instead of traditional techniques that only learn from left to right or right to left. A bi-directional architecture includes two networks, one in which the input is handled from start to end and another one from end to start. The outputs of two networks are then integrated to provide a single representation. As a result, BERT is better able to understand the relationship between words and provides better performances.


Besides, unlike other context-free word embedding models like word2vec or GloVe, which create a single word embedding representation for each word in its vocabulary, BERT needs to consider each context for a given the word in a phrase. As a consequence, homonyms in other sentences become different in each context.

In this study, we try to apply various BERT-based models to Vietnamese datasets for solving social media text classification tasks. With multilingual models, we implement multilingual BERT cased (mBERT cased), multilingual BERT uncased (mBERT uncased), XLM-RoBERTa (XLM-R), and Distil multilingual BERT (DistilmBERT). We also try the monolingual models, which are pre-trained in Vietnamese data, such as PhoBERT, viBERT, vELECTRA, and viBERT4news. Table \ref{tab:BERTology} is an overview of multilingual and monolingual BERT-based language models available in Vietnamese. All pre-trained models we used in this research are downloaded from Hugging Face\footnote{https://huggingface.co/}.

\subsection{Multilingual Language Models}
Language models are usually specifically designed and trained in English, most globally used language. Moreover, these models were then deeper trained and became multilingual to expand and serve NLP problems to support more languages globally. In this study, we deploy popular multilingual models as follows.

\subsubsection{Multilingual BERT}
\citet{devlin-etal-2019-bert} continues to develop and expand supported languages with multilingual BERT, including uncased and cased versions, after the launch of BERT. Multilingual BERT models, as opposed to their predecessors, are trained in various languages, including Vietnamese, and use masked language modeling (MLM). Each model includes a 12-layer, 768-hidden, 12-heads, and 110M parameters and supports 104 different languages. There are two multilingual BERT models: uncased\footnote{https://huggingface.co/bert-base-multilingual-uncased} and cased\footnote{https://huggingface.co/bert-base-multilingual-cased} versions, which are both used for implementation in this study.

\subsubsection{XLM-RoBERTa}
XLM-RoBERTa (XLM-R) is a cross-lingual language model provided by \citet{conneau-etal-2020-unsupervised} and trained in 100 different languages, including Vietnamese, on 2.5TB Clean CommonCrawl data\footnote{https://github.com/facebookresearch/cc\_net}. It offers significant gains in downstream tasks such as classification, sequence labeling, and question answering over previously released multilingual models like mBERT or XLM. Under XLM, the language of the input ids cannot be correctly determined by the XLM-R language for the language of the input id to be understood by language tensors. We use the XLM-R base\footnote{https://huggingface.co/xlm-roberta-base} in this research.

\subsubsection{DistilmBERT}

\begin{table*}[!htbp]
\centering
\caption{The techniques for pre-processing data we employed in experiments.}
\label{tab:preprocessing_techniques}
\begin{tabular}{clcccc} 
\hline
\multicolumn{1}{l}{\multirow{2}{*}{\textbf{No}.}} & \multicolumn{1}{c}{\multirow{2}{*}{\textbf{Pre-procesing technique}}} & \multicolumn{4}{c}{\textbf{Dataset}}  \\ 
\cline{3-6}
\multicolumn{1}{l}{}                              & \multicolumn{1}{c}{}                                                 & \textbf{VSMEC} & \textbf{ViCTSD} & \textbf{ViHSD} & \textbf{ViOCD}        \\ 
\hline
1                                                 & Removing numbers                                                     & \checkmark     &        & \checkmark     & \checkmark            \\
2                                                 & Removing punctations                                                 &       & \checkmark      & \checkmark     & \checkmark            \\
3                                                 & Removing emojis, emoticons                                           &       & \checkmark      & \checkmark     & \checkmark            \\
4                                                 & Converting emojis, emoticons into texts                              & \checkmark     &        &       &              \\
5                                                 & Tokenizing words                                                     & \checkmark     & \checkmark      & \checkmark     & \checkmark            \\
\hline
\end{tabular}
\end{table*}

DistilmBERT\footnote{https://huggingface.co/distilbert-base-multilingual-cased}, which is published by \citet{sanh2020distilbert}, is a different BERT version, in which its features improve the original version (faster, cheaper and lighter). This architecture requires less than BERT pre-training. In addition, it is more efficient compared to BERT, as the BERT model could be reduced by 40 while maintaining 97 of its language understanding capabilities and 60 more quickly. Hence, DistilmBERT has been known as a BERT version of reducing the number of parameters.

\subsection{Monolingual Language Models}
In addition to developing multilingual language models, researchers in specific languages are interested in monolingual models. Monolingual models are frequently built using the BERT architecture and pre-trained on datasets in a single language. Furthermore, because these models are trained on a large amount of data in a language, they frequently achieve great performances on NLP tasks for the languages themselves. In this study, we use several existing monolingual models that have been introduced for solving Vietnamese tasks.

\subsubsection{PhoBERT} \citet{nguyen-tuan-nguyen-2020-phobert} first presented PhoBERT models, which are pre-trained models for Vietnamese NLP. They are state-of-the-art Vietnamese language models. To handle tasks in Vietnamese, they have trained the first large-scale monolingual BERT-based with two versions as \textit{base} and \textit{large} with a 20GB word-level Vietnamese dataset combining two datasets: Vietnamese Wikipedia\footnote{https://vi.wikipedia.org/wiki/} (1GB) and modified dataset from a Vietnamese news dataset\footnote{https://github.com/binhvq/news-corpus} (19GB). The chosen version we use to implement in this study is PhoBERT base\footnote{https://huggingface.co/vinai/phobert-base}.

\subsubsection{viBERT} ViBERT\footnote{https://huggingface.co/FPTAI/vibert-base-cased}, a pre-trained language model for Vietnamese, which is based on BERT architecture and introduced by \citet{bui-etal-2020-improving}. The architecture of viBERT is similar to that of mBERT, and it has been pre-trained on large corpora of 10GB of uncompressed Vietnamese text. Nonetheless, there is a distinction between this model and mBERT. They chose to exclude insufficient vocab because the mBERT vocab still contains languages apart from Vietnamese.

\subsubsection{vELECTRA} 
ELECTRA, first introduced by \citet{clark2020electric}, is a novel pre-training architecture that uses replaced token detection (RTD) rather than language modeling (LM) or masked language modeling (MLM), as is popular in existing language models. 

\citet{bui-etal-2020-improving} also released vELECTRA, another pre-trained model for Vietnamese, along with viBERT. They used a dataset with almost 60GB of words to pre-train vELECTRA\footnote{https://huggingface.co/FPTAI/velectra-base-discriminator-cased}.

\subsubsection{viBERT4news}
NlpHUST published viBERT4news\footnote{https://github.com/bino282/bert4news}, a Vietnamese version of BERT trained on more than 20 GB of news datasets. ViBERT4news demonstrated its strength on Vietnamese NLP tasks, including sentiment analysis using comments of the AIViVN dataset, after launch and testing, with an F1 score of 0.90268 on public leaderboards (while the winner of the shared-task score is 0.90087).

\section{Experiments and Results}
In this section, we carry out experiments using monolingual and multilingual BERT-based models on Vietnamese benchmark datasets.

\begin{table*}
\centering
\caption{Experimental Results of multilingual versus monolingual models on Vietnamese social media datasets (macro-averaged F1-score (\%)).}
\label{tab:results}
\resizebox{\textwidth}{!}{%
\begin{tabular}{llcccc} 
\hline

\multicolumn{2}{c}{\textbf{Model}}                                                   & \multicolumn{1}{c}{\textbf{VSMEC}} & \multicolumn{1}{c}{\textbf{ViCTSD}} & \multicolumn{1}{c}{\textbf{ViOCD}} & \multicolumn{1}{c}{\textbf{ViHSD}}  \\ 
\hline
\multicolumn{2}{l}{\textbf{Baseline}}                                                &  \makecell{59.74 \\ \cite{uit-vsmec}}                                     &  \makecell{78.59 \\ \cite{nguyen2021constructive}}                                   &  \makecell{92.164 \\ \cite{nguyen2021vietnamese}}                                 &  \makecell{62.69 \\ \cite{luu2021largescale}}                                    \\ 
\hline
\multirow{4}{*}{\textbf{Multilingual}}                    & \textit{mBERT (cased)}   & 54.59                                  & 80.42                                   & 91.61                                & 64.20                                   \\ 
\cline{2-6}
                                                          & \textit{mBERT (uncased)} & 53.14                                  & 78.97                                   & 92.52                                & 62.76                                   \\ 
\cline{2-6}
                                                          & \textit{XLM-R}           & 62.24                                  & 80.51                                   & 94.35                                & 63.68                                   \\ 
\cline{2-6}
                                                          & \textit{DistilmBERT}     & 53.83                                  & 81.69                                   & 90.50                                 & 62.50                                   \\ 
\hline
\multicolumn{1}{c}{\multirow{4}{*}{\textbf{Monolingual}}} & \textit{PhoBERT}         & \textbf{65.44}                         & 83.55                                   & 94.71                                    & 66.07                                   \\ 
\cline{2-6}
\multicolumn{1}{c}{}                                      & \textit{viBERT}          & 60.68                                  & 81.27                                   & 94.53                                & 65.06                                   \\ 
\cline{2-6}
\multicolumn{1}{c}{}                                      & \textit{vELECTRA}        & 61.29                                  & 80.24                                   & \textbf{95.26}                                & 65.97                                   \\ 
\cline{2-6}
\multicolumn{1}{c}{}                                      & \textit{viBERT4news}     & 64.65                                  & \textbf{84.15}                          & 94.72                                    & \textbf{66.43}                          \\
\hline
\end{tabular}
}
\end{table*}

\subsection{Pre-processing Techniques}
We implement our data preprocessing strategies based on the task as well as the characteristics of the dataset. Because each dataset is distinct in terms of vocabulary, origin, and content, we use different preprocessing approaches appropriate for each dataset before feeding data into the models. Table \ref{tab:preprocessing_techniques} contains a list of techniques that we implement for each dataset.

We almost remove numbers in all tasks except the constructive speech detection task because of \citet{nguyen2021constructive} recommendation. For the VSMEC dataset, we do not remove punctuations, emojis, and emoticons because, in several cases, users often use them in their comments to express emotions, which has a great effect on the performance of the model in predicting emotion labels. Therefore, we keep the punctations, and we then apply the study of \citet{9310495} in data preprocessing, which is transforming emojis and emoticons into Vietnamese text and then obtain a higher performance with the F1-score of 64.40\% (higher than the baseline of the author of the dataset 4.66\%). Because emojis and emoticons are also essential elements influencing the emotions expressed in comments, deleting them causes a harmful effect on the emotion categorization task.

There are several preprocessing methods with different transfer learning-based models as a recommendation by its authors. For PhoBERT, we employ VnCoreNLP\footnote{https://github.com/vncorenlp/VnCoreNLP} and FAIRSeq\footnote{https://github.com/pytorch/fairseq} for preprocessing data and tokenizing words before applying them to the dataset.

\subsection{Experiment Settings}
We record and select the most parameters for each task after multiple experiment and show them in Table \ref{tab:parameters}. Additionally, we keep the max sequence length as used in the baseline of each task because they are optimized settings for it. Other parameters are remained and not customized.

\begin{table}[H]
\centering
\caption{The parameters selected for each task in the SMTCE benchmark. [1] and [2] represent two parameters batch\_size and epochs, respectively.}
\label{tab:parameters}
\resizebox{\linewidth}{!}{%
\begin{tabular}{lcccccccc} 
\hline
\multicolumn{1}{c}{} & \multicolumn{2}{c}{\textbf{VSMEC}} & \multicolumn{2}{c}{\textbf{ViOCD}} & \multicolumn{2}{c}{\textbf{ViHSD}} & \multicolumn{2}{c}{\textbf{ViCTSD}} \\ 
\hline
\multicolumn{1}{c}{} & \textit{[1]} & \textit{[2]} & \textit{[1]} & \textit{[2]} & \textit{[1]} & \textit{[2]} & \textit{[1]} & \textit{[2]} \\ 
\hline
\textit{mBERT (cased)} & 16 & 4 & 16 & 4 & 16 & 4 & 16 & 4 \\
\textit{mBERT (uncased)} & 16 & 4 & 16 & 4 & 16 & 4 & 16 & 4 \\
\textit{XLM-R} & 8 & 4 & 16 & 4 & 16 & 4 & 16 & 2 \\
\textit{DistilmBERT} & 16 & 4 & 16 & 4 & 16 & 4 & 16 & 4 \\ 
\hline
PhoBERT & 8 & 2 & 8 & 2 & 16 & 2 & 16 & 2 \\
\textit{viBERT} & 16 & 4 & 16 & 4 & 16 & 4 & 16 & 4 \\
\textit{vELECTRA} & 16 & 4 & 16 & 4 & 16 & 4 & 16 & 4 \\
\textit{viBERT4news} & 8 & 2 & 16 & 2 & 16 & 2 & 16 & 2 \\
\hline
\end{tabular}
}
\end{table}




\subsection{Evaluation Metric}
In the text classification task, we have different metrics suitable for specific datasets and problems. Because most datasets in this study are imbalanced and according to the choice of dataset authors, we choose the macro-average F1 score to evaluate the performances of models on the datasets. 

To compute the macro-average F1 score, we first calculate the F1 score per class in the dataset by the formula (1). 

\[
\text{F1 score = }\text{2}*\frac{\text{Precision * Recall}}{\text{Precision + Recall}}\quad(1)
\]

After achieving the F1 scores of all classes, we compute the macro-average F1 score by calculating the average F1 score as shown in formula (2). 

\[
\text{Macro F1 score = } \frac{sum(\text{F1 scores})}{\text{Number of classes}}\quad(2)
\]

Because the macro f1 score gives importance equally to each class, it means the macro f1 algorithm may still produce objective findings on unbalanced datasets since a majority class will participate equally alongside the minority. That is the reason why almost imbalanced dataset authors choose the macro-average f1 score as the primary metric to represent actual model performance despite skewed class sizes.

\subsection{Experimental Results}

We begin implementation after selecting appropriate parameters for each model corresponding to each task, and the results obtained by models are presented in Table \ref{tab:results}.

Our experiments outperform the original results of authors on each task in the majority of cases. As a result, we obtain the following outcomes: 65.44\% for VSMEC sentiment classification task (higher than 5.7\%) with PhoBERT model; 84.15\% for ViCTSD identifying constructiveness task (higher than 5.56\%) by viBERT4news model; 95.26\% for ViOCD classifying complaint comments task (higher than 3.1\%) by vELECTRA model; 66.43\% for ViHSD hate speech detection task (higher than 3.74\%) by viBERT4news model. The results on the ViOCD set are significantly higher than on other datasets because its data samples are processed cleanly, and the higher inter-annotator agreement during data labeling is noticeable when building the dataset.

\subsection{Result Analysis}
After achieving the results, we see that the outcomes of monolingual models are all greater than the baseline result of the author for each dataset. In terms of performance, specific techniques, like XLM-R with VSMEC; all methods with ViCTSD; mBERT cased and uncased, and XLM-R with ViHSD; and mBERT uncased and XLM-R with ViOCD, perform just above the baseline.

We also discover that the models that produce the most remarkable outcomes in this study for all tasks are monolingual. It demonstrates that monolingual language models outperform multilingual models when dealing with non-English and low-resource languages such as Vietnamese.

When we implement techniques on the VSMEC dataset, the best model is PhoBERT, which has an F1-score of 65.44\%, more significant than the result of the author of 4.66\%. \citet{huynh-etal-2020-simple} also published a method that gains an F1-score of 65.79\% (higher than ours 0.35\%). However, the approach used in that study is an ensemble classifier that combines multiple Neural Network Models such as CNN, LSTM, and mBERT. This ensemble approach must operate with a majority voting ensemble before providing the final output, which is why it takes longer to execute than a single model.



\section{Discussions on Vietnamese Social Media Text Mining}

\subsection{Monolingual versus Multilingual: Which is the better in Vietnamese social media text classification on BERTology?}

After implementing multiple multilingual and monolingual BERT-based language models for each task in the SCTE benchmark, we discovered that monolingual gets better results in most cases. Furthermore, the results of all current monolingual models outperform the baseline of the author of the dataset. Meanwhile, multilingual models only outperform baseline in a few tasks and mainly stand out with XLM-R.

Multilingual models do not now provide superior results because these models, except XLM-R, are primarily pre-trained on the Vietnamese Wikipedia corpus. This corpus is limited, with only 1GB of uncompressed data, and the material on Wikipedia is not representative of common language use. Moreover, unlike in English or other widely spoken languages, the space in Vietnamese is solely used to separate syllables, not words. Meanwhile, multilingual BERT-based models are now unaware of this. Monolingual models are pre-trained on larger and higher-quality Vietnamese datasets, such as viBERT4news on AIViVN's comments dataset or PhoBERT on a 20GB word-level Vietnamese corpus \cite{nguyen-tuan-nguyen-2020-phobert}. As a result, monolingual language models outperform bilingual language models on Vietnamese NLP tasks, particularly text classification tasks, as demonstrated by the experiments in this study. Several other tasks such as Vietnamese Aspect Category Detection \cite{dangvthin}, or Vietnamese extractive multi-document summarization \cite{to-etal-2021-monolingual} both have outperformance with monolingual BERT-based models.

Nonetheless, monolingual language models do not outperform the multilingual in all NLP tasks. More complex tasks like machine reading comprehension \cite{van2022vlsp} achieve better performances on multilingual pre-trained language models.

\subsection{How do pre-processing techniques help improve social media text mining?}

Pre-proceeding techniques are essential to improve the machine learning models significantly on social media texts, which were proven in previous studies. \citet{9310495} proposed an approach of pre-processing data before feeding data into the model in the training stage, which was also implemented in this study to pre-process the emotion recognition task. In detail, their study is to transform emojis and emoticons, which are elements to determine the whole feeling of the content but seem to be bypassed in most normal pre-processing methods, into texts to add more detail. Their methods are appropriate pre-processing techniques based on Vietnamese social media characteristics. In our experiment, the performance of the PhoBERT model improves 5.28\% from 60.16\% up to 65.44\% by using their approaches.

Additionally, carefully in the stage of building the dataset is also an effective way to gain optimum performances of models. The dataset ViOCD proves that the author worked well with strictly annotating and pre-processing data in building data phases. As a result, most models obtain excellent performance in identifying complaints on Vietnamese e-commerce websites.

Researchers have recently tended to apply various special approaches to optimize model performance. One of those methods is lexical normalization \cite{van-der-goot-etal-2021-multilexnorm}. As a characteristic of social media texts, unstructured content is an actual problem many models face. Most content on social media platforms is written in various formats and up to the users. Therefore, normalizing all of them into the standard in the pre-processing phase makes the model deeper understand the words and then achieve better performance in different natural language processing tasks.

\subsection{How do imbalanced data impact the social media text mining?}

Most datasets in social media text mining are unbalanced. Processing techniques to improve machine learning models on imbalanced datasets have recently attracted much attention from NLP researchers. As is used in this study, choosing proper metrics to evaluate models is a good method to obtain an objective view of their performances. Furthermore, if the amount of data is big enough, researchers can use the resampling method to solve the imbalance of data. The re-sampling technique reduces or extends the minority or majority class to achieve a balanced dataset. Besides traditional approaches to dealing with imbalanced data, new methods are developed to boost the model performance.  ARCID \cite{10.1007/978-3-319-73117-9_40} and EDA \cite{wei2019eda} are novel methods that can address the imbalance problem in most social media datasets. This approach seeks to extract essential knowledge from unbalanced datasets by highlighting information gathered from minor classes without significantly affecting the classifier prediction performance.
\section{Conclusion and Future Work}

This paper described a novel evaluation benchmark for social text classification named SMTCE with four tasks emotion recognition, constructive speech detection, hate speech detection, and complaint comment detection. We implemented various approaches with BERT-based multilingual versus monolingual language models on Vietnamese benchmark datasets for each SMTC task in the SMTCE benchmark. We achieved the state-of-the-art performances: 65.44\%, 84.15\%, 66.43\%, and 95.26\% for VSMEC, ViCTSD, ViHSD, and ViOCD datasets respectively.

In future, we hope that this study will serve as a standard benchmark to develop new models on Vietnamese social media text classification or mining. Furthermore, it will motivate a range of NLP benchmarks for low-resource languages like Vietnamese.

\section*{Acknowledgement}
Luan Thanh Nguyen was funded by Vingroup JSC and supported by the Master, PhD Scholarship Programme of Vingroup Innovation Foundation (VINIF), Institute of Big Data, code VINIF.2021.ThS.41.

\bibliographystyle{acl_natbib} 
\bibliography{REFERENCES}

\end{document}